\documentclass{article}

\usepackage{arxiv}

\usepackage{times}
\usepackage{soul}
\usepackage{url}
\usepackage{hyperref}
\usepackage[utf8]{inputenc}
\usepackage[small]{caption}
\usepackage{graphicx}
\usepackage{amsmath}
\usepackage{amsthm}
\usepackage{booktabs}
\usepackage{algorithm}
\usepackage{algorithmic}
\usepackage{verbatim}
\usepackage{subfigure}
\usepackage{multirow}
\usepackage{multicol}
\usepackage{natbib}

\title{An Inference Approach to Question Answering Over Knowledge Graphs}

\author{Aayushee Gupta \thanks{Work done as intern at Accenture Technology Labs in 2018.} \\
	International Institute of Information Technology, Bangalore\\
	\texttt{aayushee.gupta@iiitb.org} \\ 
	 \\
	\And
    Annervaz K M \\
	Indian Institute of Science \& Accenture Technology Labs\\
	\texttt{annervaz@iisc.ac.in} \\
	\And
	Ambedkar Dukkipati \\
	Indian Institute of Science\\
	\texttt{ambedkar@iisc.ac.in} \\
	\And
	Shubhashis Sengupta \\
	Accenture Technology Labs\\
	\texttt{shubhashis.sengupta@accenture.com} \\
}

\date{}

\begin{document}
\maketitle
\begin{abstract}

Knowledge Graphs (KG) act as a great tool for holding distilled information from large natural language text corpora. The problem of natural language querying over knowledge graphs is essential for the human consumption of this information. This problem is typically addressed by converting the natural language query to a structured query and then firing the structured query on the KG. Direct answering models over knowledge graphs in literature are very few. The query conversion models and direct models both require specific training data pertaining to the domain of the knowledge graph. In this work, we convert the problem of natural language querying over knowledge graphs to an inference problem over premise-hypothesis pairs. Using trained deep learning models for the converted proxy inferencing problem, we provide the solution for the original natural language querying problem. Our method achieves over 90\% accuracy on MetaQA dataset, beating the existing state of the art. We also propose a model for inferencing called Hierarchical Recurrent Path Encoder(HRPE). The inferencing models can be fine tuned to be used across domains with less training data. Our approach does not require large domain specific training data for querying on new knowledge graphs from different domains.

\end{abstract}

\section{Introduction \& Motivation}

\begin{figure*}
\begin{center}
\includegraphics[width=0.9\textwidth,height=0.25\textheight]{{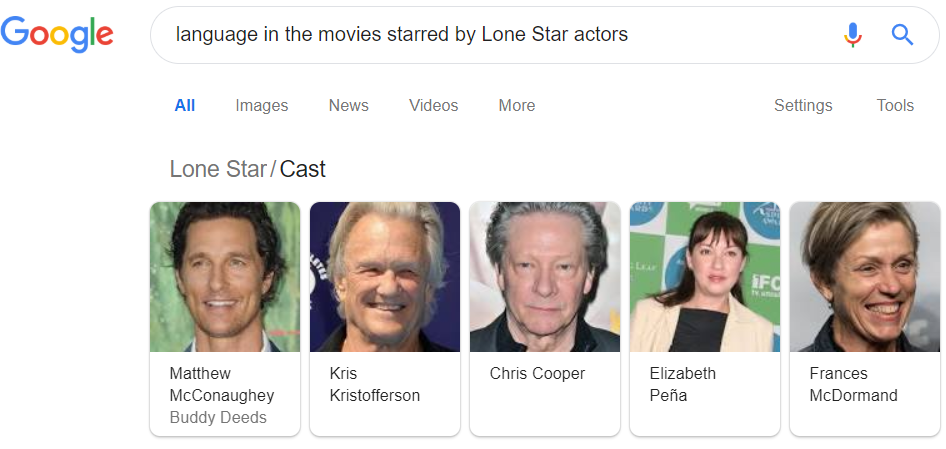}}
\caption{Complex natural language query search results on Google}
\label{fig:google}
\end{center}
\end{figure*}

 Knowledge Graphs ~\citep{nickel2016review} \footnote{https://googleblog.blogspot.in/2012/05/introducing-knowledge-graph-things-not.html}
 represent information in the form of fact triplets, consisting of a subject entity, relation and an object entity (example: $<$\textit{Italy, capital, Rome}$>$). The entities represent the nodes of the graph and the relationships between them act as edges. A fact triple $($subject entity, relation, object relation$)$ is represented as $(h,r,t)$.  Practical knowledge graphs congregate information from secondary databases or extract facts from unstructured text using various statistical learning mechanisms, examples of such a system is NELL~\citep{mitchell2015never}. There are human created knowledge bases as well, like Freebase (FB15k)~\citep{bollacker2008freebase}. Knowledge graphs are a great tool to hold distilled information for human consumption as well as using this information for machine processing. For human consumption, natural language querying capability over these graphs is quite essential.
 A natural language query can vary from very simple forms like \textit{what is the capital of Italy?}, which can be answered from a single fact, to complex forms which needs to be reasoned over a set of connected facts. Figure~\ref{fig:google} shows an example of a complex query unanswered by Google Knowledge Graph. The query requires reasoning based on multiple facts collected from the knowledge graph involving the query and answer entities.

We approach the problem of natural language querying from the perspective of inference. The main motivation is that a Knowledge Graph contains facts to support a given answer. So the facts from KG acting as a premise, should be able to tell whether an answer is right or wrong. Also, there has been considerable work done in the domain of inferencing, with multitudes of datasets. If we convert the original problem to inferencing, we could leverage that volume of work. The problem of Natural Language Inference (NLI) is to identify whether a statement (hypothesis: $\mathcal{H}$) in natural language can be supported or contradicted in the context of another statement (premise: $\mathcal{P}$) in natural language. If it can neither be inferred nor contradicted, we say the hypothesis is `neutral' to the premise. NLI is one of the most important components for natural language understanding systems~\citep{benthem2008brief,maccartney2009extended}. For the original problem, given the query we create hypothesis facts based on potential answers to the query and populate a set of premises for each based on the facts related to them in the KG. For the earlier example, two potential hypotheses are \textit{Milan is the capital of Italy} and \textit{Rome is the capital of Italy}. Separate premises for each hypothesis are created based on various facts available in KG related to \textit{Italy, Milan} and \textit{Italy, Rome}. These premise-hypothesis pairs are tested for entailment or contradiction using the inference models. The above is a simple example, the premises and hypotheses get increasingly long and complex based on the complexity of the query.

Some of the state of the art models for NLI are ESIM (Enhanced LSTM model)\citep{chen2016enhanced,lan2018neural} and Transformer\citep{radford2018improving}.  We used ESIM model in our work. It is a combination of the Bi-LSTM~\citep{greff2015lstm} and Tree LSTM~\citep{tai2015improved} models which encode the premises and hypotheses sentences and their parse trees, respectively. Further, local inferences for words and their context, along with local information between (linguistic) phrases and clauses are collected through an attention model~\citep{luong2015effective,bahdanau2014neural}.

The main contributions of our work are as follows,
\begin{enumerate}
	\item An approach to convert \textit{Question Answering}(QA) problem over a Knowledge Graph to that of \textit{Inference}.
	\item Hierarchical Recurrent Path Encoder(HRPE), which is a custom model for the inference problem at hand.
	\item Using the approach and leveraging the work on Natural Language Inference, we achieved state of the art accuracy on MetaQA~\citep{zhang2018variational} dataset.
	\item The proposed approach is amenable to domain adaptation, and this property can be utilized for Knowledge Graphs from new domains with fewer training data.
\end{enumerate}

\section{Our Approach and Model}
\label{sec:main}

For the original problem, given a Knowledge Graph $\mathcal{K}$ and a natural language query $Q$ with entities mentioned $\mathcal{Q}_e = \{e_1,\dots,e_n\}$, the problem is to retrieve the correct set of answer entities $\mathcal{A}_c$ from nodes of $\mathcal{K}$. We first populate a potential answer set $\mathcal{A}_p$ from the connected entities of $Q_e$ in $\mathcal{K}$. For each $a_i \in \mathcal{A}_p $, the various paths in  $\mathcal{K}$ from $a_i$ to each element of $\mathcal{Q}_e$ is populated and considered a part of the premise $\mathcal{P}$. Similarly, for each $a_i \in \mathcal{A}_p$, WH-word in the query is replaced with $a_i$ to form the hypothesis $\mathcal{H}_i$. The premise-hypothesis pair $(\mathcal{P},\mathcal{H}_i)$, is then processed using inferencing models to check for Entailment and Contradiction. Entailment implies that the answer $a_i$ is in the correct answer set $\mathcal{A}_c$, else discarded as wrong answer. The set $\mathcal{A}_c$ is finally given as the result for the original natural language query.

\begin{figure*}
\begin{center}
\includegraphics[width=0.8\textwidth,height=0.3\textheight]{{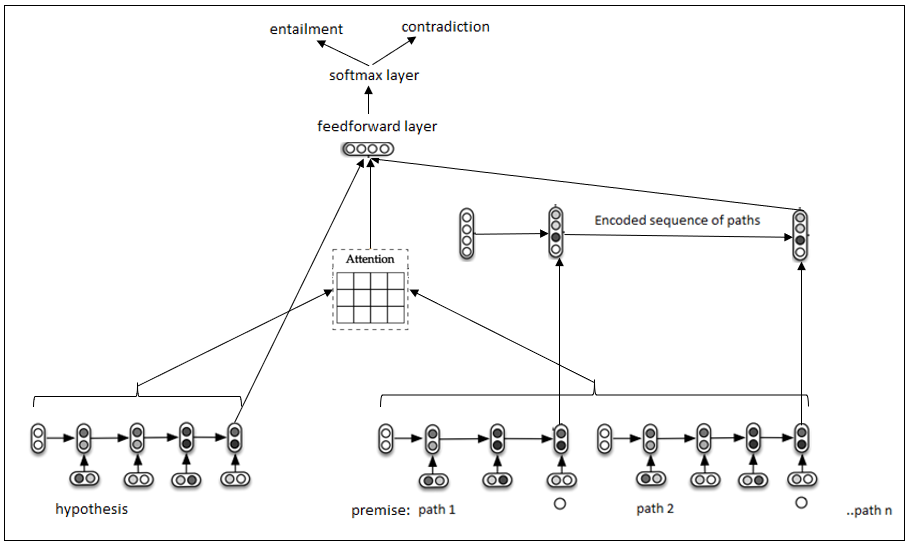}}
\caption{Hierarchical Recurrent Path Encoder Model(HRPE)}
\label{fig:path}
\end{center}
\end{figure*}

A path between two entities $e_i$ and $e_j$ in $\mathcal{K}$ is an \textit{ordered set} of triples $\mathcal{P}_{ij}$. If they are directly connected, then $P_{ij}$ contains only a single triple, $\{(e_i,r_{ij},e_j)\}$, where $r_{ij}$ stands for the relationship between $e_i$ and $e_j$. If the length of the path connecting them is $n$, then $P_{ij}$ will contain $n$ such triples. Since there can be multiple paths between two entities in $\mathcal{K}$, all paths upper bounded by a hyper parameter for length are considered in the premise path set, $\mathcal{P}$.  After forming the set of connected paths that form the premise, we process the premise in two different ways. 

Each fact(a single triplet in single path) in the premise is converted to natural language form using templates. For example \textit{(Donald Trump,presidentOf,USA)} is converted to \textit{Donald Trump is the president of the USA}. These type of templates can be easily generated for relationship types in $\mathcal{K}$. In this method, the premise is a set of sentences and hypothesis is a single sentence. This premise-hypothesis pairs dataset is processed using ESIM(Enhanced LSTM model)\citep{chen2016enhanced,lan2018neural} model akin to standard natural language inference. For the entities and relationship tokens, the TransE~\citep{bordes2013translating} embeddings of $\mathcal{K}$ are used and for the other template tokens, pre-trained GloVe~\citep{pennington2014glove} embeddings were used. Our model does not assume any specific KG embeddings or its properties. For the current work, we have used the simplest of the lot (TransE). We use the TransE model implementation from OpenKE\citep{han2018openke} to generate 300 dimensional embeddings for entities and relations in the KG.

\paragraph{Hierarchical Recurrent Path Encoder Model(HRPE):}
The conversion of premise path triplets was done to align the problem to that of natural language inference. But we realized that may not be the best approach to create a good representation for the premise, for one the NLI models were discarding the granularity of a single path, which is important in the final inference. Also the sequence lengths were getting large, forcing the model to discard many samples from processing. Hence we came up with Hierarchical Recurrent Path Encoder Model(HRPE). We first encoded each path $P_j \in \mathcal{P}$ using an LSTM, equation~\ref{eq:path}. The hypothesis $\mathcal{H}_i$ was encoded to $H$, using a bi-LSTM, equation~\ref{eq:hypo}. This was used to generate attention over each path encodings, and creating a single path encoding($p$) based on the attention weights, equation~\ref{eq:path2}. This allowed the model to focus on the premise component important from the hypothesis perspective,i.e., the correct path between query and answer entity. The path encodings were also further encoded by another LSTM, in hierarchical fashion to generate $P$, equation~\ref{eq:path2}. This was required to allow interaction between paths for the final inference. The concatenated vector $H:p:P$ was fed through a Feed Forward network to do the classification prediction, equations~\ref{eq:class},~\ref{eq:loss}. 

\begin{equation}
\forall  P_j \in \mathcal{P},\  p_j = (p_{j1}, \cdots, p_{jn}) \leftarrow \mathrm{LSTM}(P_j)
\label{eq:path}
\end{equation}

\begin{equation}
H = (h_{i1}, \cdots, h_{in}) \leftarrow \mathrm{bi-LSTM}(\mathcal{H}_i)
\label{eq:hypo}
\end{equation}

\begin{equation}
p = \Sigma_j \frac{H.(p_j)^T}{\Sigma_k H.(p_k)^T} (p_{j1}, \cdots, p_{jn})
\label{eq:path2}
\end{equation}

\begin{equation}
P = (P_1, \cdots, P_n) \leftarrow \mathrm{LSTM}(p_1,\cdots,p_j)
\label{eq:path3}
\end{equation}

The aggregated vector $\mathcal{A} = H:p:P$ was then used to do the
final label prediction of entailment or contradiction, 
\begin{equation}
\mathrm{label} = \mathrm{softmax}(W^T \mathcal{A} )),
\label{eq:class}
\end{equation}
and
\begin{equation}
\mathrm{loss} = C(\mathrm{label}_{\mathrm{gold}}, \mathrm{label}),
\label{eq:loss}
\end{equation}
    where $W$ is the model parameter and $C(p,q)$ denotes the
    cross-entropy between $p$ and $q$. We minimized this loss averaged
    across the training samples, to learn the various model parameters
    using Stochastic Gradient Descent~\citep{stochastic-gradient-tricks} algorithm. 

\begin{figure*}
\includegraphics[width=\textwidth,height=0.4\textheight]{{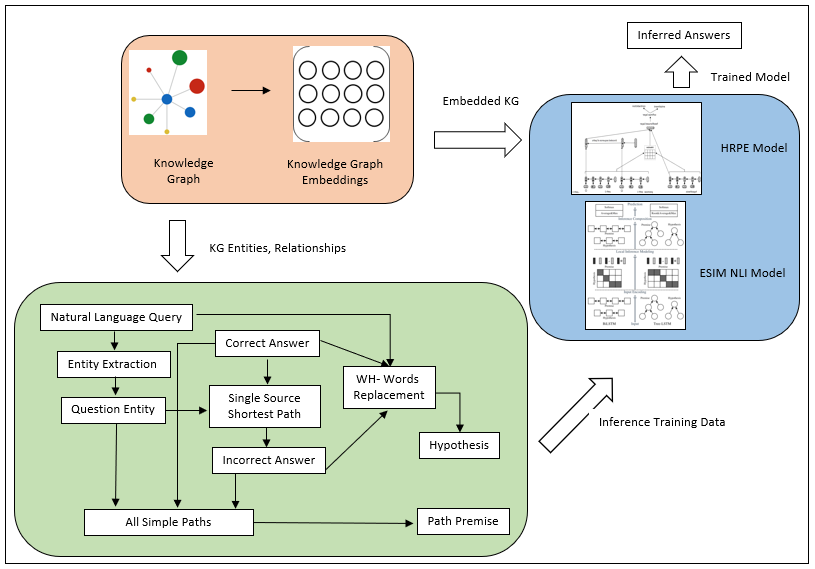}}
\caption{Proposed solution framework for the problem}
\label{fig:framework}
\end{figure*}

Our complete solution framework consists of several components for working with knowledge graphs, extraction and linking of entities between queries and knowledge graphs, converting queries to a multiple choice query-answer format followed by conversion to premise-hypothesis-label format and finally training the inference models on this data to predict query answers. The whole framework and dependency between components is shown in Figure~\ref{fig:framework}.

\subsection{Domain Adaptability of the Model}

In our approach and proposed model, two kinds of tokens are input to the model, the entity-relationship tokens and template/question word tokens. We used generated TransE~\citep{bordes2013translating} KG embeddings for entities and relationship tokens while pre-trained Glove ~\citep{pennington2014glove} 300D embeddings for rest of the words in the vocabulary. The template/question word tokens are very few in number compared to entity relationship tokens. The KG embeddings can be generated in an unsupervised fashion given a new Knowledge Graph. The concept of inferencing has less gravity on the domain and the learnings are thus transferable across domains. Hence, we postulate that our approach can be used to create QA models for domains where KG is available, but have fewer QA training data. As such, the models trained on one source domain QA training data $\mathcal{D}_s$, can be used with minimal fine tuning on new target domain $\mathcal{D}_t$. A transformation of the KG embeddings between source and target domain can also be learned in an unsupervised manner and can be fine tuned with fewer training data from the target domain, equation~\ref{eq:domain1}. This will help in better results for domain adaptation.

\begin{equation}
 f( W, E(\mathcal{K}_s)) = E(\mathcal{K}_t)
 \label{eq:domain1}
\end{equation}

Here, $W$ are the model parameters for the transformation function $f$(learned unsupervised and then fine tuned), $E$ is the KG embeddings generation function and $\mathcal{K}_s, \mathcal{K}_t$ are source and target domain KGs respectively.

\begin{equation}
\mathcal{I}_s \leftarrow \mathrm{Train}(\mathcal{D}_s,E(\mathcal{K}_s)))
\end{equation}

Here, $\mathcal{I}_s$ is the trained inference model from the source domain. For creating the model on the target domain $\mathcal{I}_t$, we use the trained model from the source domain, learned transformation function for KG embeddings and fine tuned training data from target domain.

\begin{equation}
\mathcal{I}_t \leftarrow \mathrm{Train}(\mathcal{D}_t,f(W,E(\mathcal{K}_s))) : \mathcal{I}_s
\end{equation}

\section{Experiments \& Results}
\label{sec:exp}

\paragraph{Dataset and Preprocessing:}
\begin{itemize}
    
\item MetaQA

We used MetaQA dataset~\citep{zhang2018variational} anchored over a movie knowledge graph having complex queries. The dataset contains 400K QA pairs, 100k of which can be answered using single path length, 150K require paths of lengths 2 in premise and 150K require paths of length 3 in premise.The anchored WikiMovies\footnote{https://research.fb.com/downloads/babi/} knowledge graph contains over 1 million triples having 38340 entities and 6 relationship types covering directors,writers,actors,languages,release years and languages associated with movies. For each query, we populated a total of $n$ answer choices including the set of correct and incorrect answers. The premise-hypothesis pairs were then labelled as entailment for the correct answers and contradiction for the wrong answers. An illustrative example of the generated PHL triplet is given in Table \ref{table:example}. Premise-Hypothesis-Label(PHL) triplet data was then further split to test and train the inference models. Certain questions were discarded from the original dataset, since the sequence length for processing premise was exorbitantly high. The dataset statistics after cleaning up are given in the Table~\ref{table:stats}. Entity identification from query was done using bi-LSTM-CRF model\citep{lample2016neural}. The identified entities were linked to the entities in the knowledge graph using Jaro Winkler~\citep{deza2009encyclopedia}similarity measure. 
\item PathQuestions

PathQuestions (PQ)~\citep{zhou2018interpretable} is another smaller multi-hop question answering dataset developed from a subset of Freebase Knowledge Graph of 2 million triples having 2215 entities and 14 relationships. Each path question consists of natural language question related to a topic entity whose answer can be found in the KG by following the correct answer path. The PQ dataset contains a total of 1908 and 5198 questions for paths of lengths 2 and 3 respectively which we use for our experiments. We generate the set of correct and incorrect answers followed by creation of PHL triplets in the same way as mentioned above. The dataset statistics for the same are shown in Table \ref{table:stats}. We follow the same dataset split as mentioned in ~\citep{zhou2018interpretable}.

\end{itemize}

\paragraph{Implementation, Training \& Hyper-parameters:}   The model was implemented in TensorFlow~\citep{tensorflow2015-whitepaper} - an open-source library for numerical computation for Deep Learning. All experiments were carried on a Dell Precision Tower 7910 server with Nvidia Titan X GPU. The models were trained using the Adam's Optimizer~\citep{kingma2014adam} in a stochastic gradient descent~\citep{stochastic-gradient-tricks} fashion. We used batch normalization~\citep{ioffe2015batch} while training and dropout for regularization. The ESIM model gave best results when hidden layer of LSTM was of size 300, equivalent to the word embedding dimension. On increasing the sequence length for ESIM model, the model became too heavy and was unable to process all the questions in the 3-length-path dataset having long premises. For HRPE model, the LSTM hidden state size of 150 gave the best performance and a sequence length of 20 and 250 were used in the hierarchical levels of LSTM.

 \paragraph{Metrics:} The classification accuracy of each model was calculated as usual based on the  predicted class and the gold label for each PHL sample. For QA accuracy, the set of predicted answers are populated based on the inference prediction with the hypothesis created from potential answer set. The aggregated correct answers for each question are matched with the gold labeled answers for each query and if both the sets match, then a question is considered to have been answered as correct, otherwise incorrect.

\subsection{Results}

The results for the experiments with the two inference models for 2 and 3 length path QA pairs are presented in Table \ref{table:res}. We have obtained state of the art accuracy on both 2 and 3 path length QA with both ESIM and HRPE models, beating the variational model(VRN)\citep{zhang2018variational} published along with the MetaQA dataset.Table \ref{table:res} presents accuracy calculated via Hit@1 metric used in the VRN model while we use an Exact Answer Set Match metric for the QA accuracy calculation. Our models clearly perform better, especially in the 3 length path case. However with the smaller PQ dataset, we have less accuracy compared to the baseline IRN \citep{zhou2018interpretable} model for 2-length path while our HRPE model performs better in the 3-length path case.
We also present domain adaptability results in Table \ref{table:domain} by pre-training both HRPE and ESIM models with (source domain) MetaQA dataset and fine-tune them on the PQ dataset (target domain). The results indicate the feasibility of our approach across different domain KGs although the ESIM model is better suited to domain adaptability compared to the HRPE model. 


\begin{table}
\centering
\begin{tabular}{|l|l|l|} 
\hline
\multicolumn{3}{|l|}{Question: Which person directed the films acted by the actors in Kid Millions?}       \\ 
\hline
\multicolumn{3}{|l|}{Potential Answers: David Butler, Douglas Sirk, Tom Hooper, Franck Khalfoun}                \\ 
\hline
Premise  & Hypothesis     & Label  \\ 
\hline
\begin{tabular}[c]{@{}l@{}}Kid Millions starred actors Eddie Cantor and \\ Thank Your Lucky Stars starred actors Eddie Cantor and \\ Thank Your Lucky Stars directed by David Butler\\ \end{tabular} & \begin{tabular}[c]{@{}l@{}}David Butler directed the films \\acted by the actors in Kid Millions\end{tabular}     & 0      \\ 
\hline
\begin{tabular}[c]{@{}l@{}}Kid Millions release year 1934 and\\Imitation of Life release year 1934 and\\Imitation of Life directed by Douglas Sirk\end{tabular}                                & \begin{tabular}[c]{@{}l@{}}Douglas Sirk directed the films \\acted by the actors in Kid Millions \end{tabular}    & 1      \\ 
\hline
\begin{tabular}[c]{@{}l@{}}Kid Millions release year 1934 and\\Les Miserables release year 1934 and\\Les Miserables directed by Tom Hooper\end{tabular}                                      & \begin{tabular}[c]{@{}l@{}}Tom Hooper directed the films \\acted by the actors in Kid Millions \end{tabular}      & 1      \\ 
\hline
\begin{tabular}[c]{@{}l@{}}Kid Millions release year 1934 and \\ Maniac release year 1934 and \\ Maniac directed by Franck Khalfoun\\ \end{tabular}                                                  & \begin{tabular}[c]{@{}l@{}}Franck Khalfoun directed the films \\acted by the actors in Kid Millions \end{tabular} & 1      \\
\hline
\end{tabular}
\caption{Sample example query conversion to NLI format: Named entity in the question (Kid Millions) is linked to the correct answer entity generating an Entail (label 0) PHL triplet while incorrect potential answers found at n-path length away from the question entity in the KG are used to generate Contradiction (label 1) PHL triplet.}
\label{table:example}
\end{table}

\begin{table}
\centering

\begin{tabular}{|l|l|l|l|l|l|l|l|} 
\hline
\multicolumn{2}{|l|}{Dataset Type}                                                                                       & \multicolumn{3}{l|}{MetaQA}                                   & \multicolumn{3}{l|}{PQ}                                         \\ 
\hline
\begin{tabular}[c]{@{}l@{}}Path length\\\end{tabular} & Model                                                            & ESIM  & HRPE  & \begin{tabular}[c]{@{}l@{}}VRN \end{tabular} & ESIM  & HRPE   & \begin{tabular}[c]{@{}l@{}}IRN \end{tabular}  \\ 
\hline
\multirow{3}{1cm}{two length path}           &\begin{tabular}[c]{@{}l@{}}Classification\\Accuracy\end{tabular} & 99.8  & 98.4  & {-}                                             & 73.3  & 77.75  & {-}                                              \\ 
\cline{2-8}
                                                      & \begin{tabular}[c]{@{}l@{}}QA\\Accuracy\end{tabular}             & 99.29 & 95    & 91.9                                          & 51.31 & ~56.54 & 96                                             \\ 
\cline{2-8}
                                                      & \#Test QA                                                        & 10045 & 10045 & 10045                                         & 191   & 191~   & 191                                            \\ 
\hline
\multirow{3}{1cm}{three length path}                        & \begin{tabular}[c]{@{}l@{}}Classification\\ Accuracy\end{tabular} & 97.1  & 97.71 & -                                             & 92.27 & 94.18  & -                                              \\ 
\cline{2-8}
                                                      & \begin{tabular}[c]{@{}l@{}}QA\\Accuracy\end{tabular}             & 89.43 & 84.77 & 58/59.7                                       & 84.93 & 88.96  & 87.7                                           \\ 
\cline{2-8}
                                                      & \#Test QA                                                        & 2665  & 4969  & 2665/4969                                     & 272   & 335    & 520                                            \\
\hline
\end{tabular}
\caption{Results from inference models along with benchmark models comparison}
\label{table:res}
\end{table}


\begin{table}
\centering
\begin{tabular}{|l|l|l|l|l|l|} 
\hline
\multicolumn{2}{|l|}{\textbf{Dataset }}                                                                                            & \multicolumn{2}{l|}{\textbf{MetaQA }}                                                                            & \multicolumn{2}{l|}{\textbf{PQ}}                                                                       \\ 
\hline
\begin{tabular}[c]{@{}l@{}}\textbf{Question}\\\textbf{ Type} \end{tabular} &                                                       & \textbf{Questions}                                     & \textbf{PHLs}                                           & \textbf{Questions}                                & \textbf{PHLs}                                      \\ 
\hline
2-length                                                                   & \begin{tabular}[c]{@{}l@{}}Train\\ Test \end{tabular} & \begin{tabular}[c]{@{}l@{}}80024\\ 10045 \end{tabular} & \begin{tabular}[c]{@{}l@{}}320096\\ 40180 \end{tabular} & \begin{tabular}[c]{@{}l@{}}1526\\191\end{tabular} & \begin{tabular}[c]{@{}l@{}}3052\\382\end{tabular}  \\ 
\hline
3-length                                                                   & \begin{tabular}[c]{@{}l@{}}Train\\ Test \end{tabular} & \begin{tabular}[c]{@{}l@{}}41575\\ 4969 \end{tabular}  & \begin{tabular}[c]{@{}l@{}}166300\\ 19876 \end{tabular} & \begin{tabular}[c]{@{}l@{}}2730\\335\end{tabular} & \begin{tabular}[c]{@{}l@{}}5460\\670\end{tabular}  \\
\hline
\end{tabular}
\caption{Data statistics after conversion from MetaQA and PQ}
\label{table:stats}

\end{table}

\begin{table}
\centering
\begin{tabular}{|l|l|l|l|l|l|} 
\hline
\multirow{2}{*}{Path Length} & \multirow{2}{*}{Model} & \multicolumn{2}{l|}{\begin{tabular}[c]{@{}l@{}}Without Domain\\Adaptation \end{tabular}}                                & \multicolumn{2}{l|}{\begin{tabular}[c]{@{}l@{}}With Domain\\Adaptation \end{tabular}}                                    \\ 
\cline{3-6}
                             &                        & \begin{tabular}[c]{@{}l@{}}Classification\\Accuracy\end{tabular} & \begin{tabular}[c]{@{}l@{}}QA\\Accuracy\end{tabular} & \begin{tabular}[c]{@{}l@{}}Classification\\Accuracy\end{tabular} & \begin{tabular}[c]{@{}l@{}}QA\\Accuracy\end{tabular}  \\ 
\hline
\multirow{2}{*}{2-length}    & ESIM                   & 73.3                                                             & 51.31                                                & 76.7                                                             & 58.64                                                 \\ 
\cline{2-6}
                             & HRPE                   & 77.75                                                            & 56.54                                                & 78.27                                                            & 56.02                                                 \\ 
\hline
\multirow{2}{*}{3-length}    & ESIM                   & 92.27                                                            & 84.93                                                & 93.9                                                             & 87.87                                                 \\ 
\cline{2-6}
                             & HRPE                   & 94.18                                                            & 88.96                                                & 94.33                                                            & 88.66                                                 \\
\hline
\end{tabular}
\caption{Results from Domain Adaptation on PQ dataset}
\label{table:domain}
\end{table}

\begin{figure*}
    \centering
    \subfigure[Classification Accuracy vs Number of Answers]{{\includegraphics[width=0.45\textwidth]{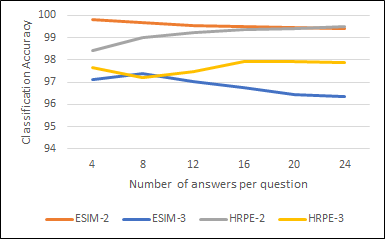} }}%
    \hfill
    \subfigure[QA Accuracy vs Number of Answers]{{\includegraphics[width=0.45\textwidth]{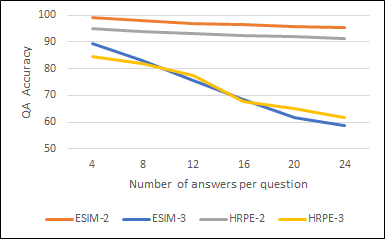} }}%
    \hfill
    \caption{Effect on model accuracy (classification and QA) on varying Number of potential answers per question for 2 and 3 length path QA. ESIM-2 and ESIM-3 refer to the NLI model for 2 and 3 path length QA while HRPE-2 and HRPE-3 refer to the path encoder model}%
    \label{fig:ablation}
\end{figure*}

As variational model(VRN)\citep{zhang2018variational} reported Hit@1 metric, we performed a model ablation study by modifying the number of potential answers considered per question. We vary the potential answer choices from 4 to 24 to see the effect on inference models accuracy  Figures~\ref{fig:ablation}. Classification accuracy of the inference models is not affected much but we observe a  drop in QA accuracy in case of 3 length path case while the 2 length path cases still performs better. Also, the HRPE model can be seen less inclined towards loss in accuracy compared to the ESIM model. Even at 24 answers considered, our numbers are better than VRN model. Our model can discard more wrong answers effectively, as wrong answer entities have lesser chance of good premise connectivity to the entities in the question.

\section{Related Work}
\label{sec:related}

\paragraph{Inferencing over Knowledge Graphs:} Typically in literature, Inferencing over Knowledge Graphs is an entirely different problem. Knowledge Graphs are largely constructed from text using statistical techniques ~\citep{mitchell2015never,niu2012deepdive}. Such Knowledge Graphs are generally incomplete. Inferencing over KGs generally refers to predicting the missing links in these incomplete KGs, also referred in literature as KG Completion solved using Embedding and Path ranking based models \citep{xiong2017deeppath,das2017go,chen2018variational}.
These methods perform inference using probabilistic or reinforcement learning models trained over existing KG triples and predict links missing from the graph by going over similar paths and creating new relations for non-neighbor entity pairs. Our problem is entirely different, we assume the KG is complete and then do Question Answering on KG, reasoned over multiple facts available.

\paragraph{Question Answering over Knowledge Graphs:} One major approach for natural language querying on Knowledge Graphs or other structured sources is converting the query into a structured query language and firing it on the KG. Existing models use a parse tree to convert query to a structured form ~\citep{liang2016learning}. Latest works treat this as a machine translation problem, with source language as the natural language and target language as the structured query language~\citep{wang2015building},~\citep{zhong2017seq2sql}, ~\citep{xu2017sqlnet},~\citep{mccann2018natural}. Our approach is different, since we do not convert the query into intermediate structured form and generate the answer directly from the KG. The closest work in literature, with respect to what we proposed here is Variational Reasoning Network\citep{zhang2018variational}. It is a probabilistic inference model for answering movie based questions that require reasoning over KG triples for answering them. We use the dataset published in this work and their results as benchmark. However, our approach and model is completely different from this work. The PathQuestions~\citep{zhou2018interpretable} dataset that we use in our experiments had a benchmark IRN model wherein annotations are used for every question with correct relation chain or path, topic and answer entities already known while we derive the paths between question and answer entity from the KG in our approach.  Question Answering over text has been attempted from inference perspective in works like KG2~\citep{zhang2018kg}. However this is being done over text, and their model uses KG created from text hypothesis and premise to do the inferencing. In contrast, we are doing Question Answering over KG and our model is completely different.

There are plethora of question answering datasets available in which questions are answered by retrieving information from a corpus. We particularly worked with complex natural language queries in our study. Such non-factoid questions are not easily answered by a retrieval engine and require some reasoning model that can combine multiple facts from a KG leading to the correct answers.~\citep{talmor2018web} provide a set of complex web questions based on the Freebase KG and a seq-to-seq model to obtain answers to these questions via decomposition to simple question answers assuming the simple questions can be answered via a retrieval engine. WikiHop \citep{welbl2018constructing} and HotPotQA\citep{yang2018hotpotqa} are based on questions which can be answered by reading multiple Wikipedia pages. These datasets follow a reading comprehension based question answering task requiring reasoning over multiple text documents. ARC \citep{clark2018think} is another multiple-choice science based question answering dataset which requires complex reasoning from scientific knowledge text corpus to answer questions prepared with the help of school instructors of grades 6-8. Scitail\citep{khot2018scitail} is an entailment dataset developed from Science based QA and has been shown to be useful for answering the complex multiple choice ARC science questions. Most of these works are  based on reasoning or retrieval over a text corpus.\footnote{Our work stopped in late 2019. And relevant literature post that is not considered or cited here.}

\section{Conclusion \& Future Work}
\label{sec:future}

We presented a simple approach for converting Question Answering over Knowledge Graphs into an inference problem. Leveraging existing models of Natural Language Inference, as well as proposing a new model, we have shown state of the art results on MetaQA dataset. Our model is simple and amenable for domain adaptation, to solve the problem of QA over KGs from newer domains with lesser training data. To the best of our knowledge, this is the first attempt in treating QA over KG as an inferencing problem and the results are exciting. The work is preliminary and provides a good starting point for discussion and further research. 

In the related work, we mentioned about KG completion problem and there is a considerable volume of work in that space. Although we positioned the work as QA reasoned over known facts from the KG, QA over KG can involve reasoning over unknown facts as well. We can leverage the models for KG completion and attempt to solve the problem of QA over KG reasoned over both known and unknown facts. The problem area is relatively new, and datasets require correspondence between the QA training data and the Knowledge Graph. There are not many datasets available to quantify the results of domain adaptation yet. We are currently working on creating a new dataset for KG question answering based on ComplexWebQA dataset~\citep{talmor2018web} which has Freebase~\citep{bollacker2008freebase} as the anchored KG. The creation of this dataset is essential to test the model capabilities to reason over larger fact sets since the existing MetaQA dataset is simple in that regard. We have considered the potential answers(correct and incorrect) from the model and generated them in the preprocessing step. Inclusion of generating potential answers using a mechanism such as attention over the graph, within the model, is an adaptation that can be taken up. We hope our work and initial results will inspire the community to conduct further investigation in this area.

\newpage
\bibliographystyle{acl_natbib}
\bibliography{IAQAKG}

\begin{thebibliography}{38}
\expandafter\ifx\csname natexlab\endcsname\relax\def\natexlab#1{#1}\fi

\bibitem[{Abadi et~al.(2015)Abadi, Agarwal, Barham, Brevdo, Chen, and
  et~al.}]{tensorflow2015-whitepaper}
Mart\'{\i}n Abadi, Ashish Agarwal, Paul Barham, Eugene Brevdo, Zhifeng Chen,
  and Craig~Citro et~al. 2015.
\newblock \href {http://tensorflow.org/} {{TensorFlow}: Large-scale machine
  learning on heterogeneous systems}.
\newblock Software available from tensorflow.org.

\bibitem[{Bahdanau et~al.(2014)Bahdanau, Cho, and Bengio}]{bahdanau2014neural}
Dzmitry Bahdanau, Kyunghyun Cho, and Yoshua Bengio. 2014.
\newblock Neural machine translation by jointly learning to align and
  translate.
\newblock \emph{arXiv preprint arXiv:1409.0473}.

\bibitem[{Benthem(2008)}]{benthem2008brief}
Johan~van Benthem. 2008.
\newblock \emph{A brief history of natural logic}.
\newblock College Publications.

\bibitem[{Bollacker et~al.(2008)Bollacker, Evans, Paritosh, Sturge, and
  Taylor}]{bollacker2008freebase}
Kurt Bollacker, Colin Evans, Praveen Paritosh, Tim Sturge, and Jamie Taylor.
  2008.
\newblock Freebase: a collaboratively created graph database for structuring
  human knowledge.
\newblock In \emph{Proceedings of the 2008 ACM SIGMOD international conference
  on Management of data}, pages 1247--1250. AcM.

\bibitem[{Bordes et~al.(2013)Bordes, Usunier, Garcia-Duran, Weston, and
  Yakhnenko}]{bordes2013translating}
Antoine Bordes, Nicolas Usunier, Alberto Garcia-Duran, Jason Weston, and Oksana
  Yakhnenko. 2013.
\newblock Translating embeddings for modeling multi-relational data.
\newblock In \emph{Advances in neural information processing systems}, pages
  2787--2795.

\bibitem[{Bottou(2012)}]{stochastic-gradient-tricks}
L\'{e}on Bottou. 2012.
\newblock \href
  {https://www.microsoft.com/en-us/research/publication/stochastic-gradient-tricks/}
  {\emph{Stochastic Gradient Tricks}}, volume 7700, page 430â€“445.
  Springer.

\bibitem[{Chen et~al.(2016)Chen, Zhu, Ling, Wei, Jiang, and
  Inkpen}]{chen2016enhanced}
Qian Chen, Xiaodan Zhu, Zhenhua Ling, Si~Wei, Hui Jiang, and Diana Inkpen.
  2016.
\newblock Enhanced lstm for natural language inference.
\newblock \emph{arXiv preprint arXiv:1609.06038}.

\bibitem[{Chen et~al.(2018)Chen, Xiong, Yan, and Wang}]{chen2018variational}
Wenhu Chen, Wenhan Xiong, Xifeng Yan, and William Wang. 2018.
\newblock Variational knowledge graph reasoning.
\newblock \emph{arXiv preprint arXiv:1803.06581}.

\bibitem[{Clark et~al.(2018)Clark, Cowhey, Etzioni, Khot, Sabharwal, Schoenick,
  and Tafjord}]{clark2018think}
Peter Clark, Isaac Cowhey, Oren Etzioni, Tushar Khot, Ashish Sabharwal, Carissa
  Schoenick, and Oyvind Tafjord. 2018.
\newblock Think you have solved question answering? try arc, the ai2 reasoning
  challenge.
\newblock \emph{arXiv preprint arXiv:1803.05457}.

\bibitem[{Das et~al.(2017)Das, Dhuliawala, Zaheer, Vilnis, Durugkar,
  Krishnamurthy, Smola, and McCallum}]{das2017go}
Rajarshi Das, Shehzaad Dhuliawala, Manzil Zaheer, Luke Vilnis, Ishan Durugkar,
  Akshay Krishnamurthy, Alex Smola, and Andrew McCallum. 2017.
\newblock Go for a walk and arrive at the answer: Reasoning over paths in
  knowledge bases using reinforcement learning.
\newblock \emph{arXiv preprint arXiv:1711.05851}.

\bibitem[{Deza and Deza(2009)}]{deza2009encyclopedia}
Michel~Marie Deza and Elena Deza. 2009.
\newblock Encyclopedia of distances.
\newblock In \emph{Encyclopedia of Distances}, pages 1--583. Springer.

\bibitem[{Greff et~al.(2015)Greff, Srivastava, Koutn{\'\i}k, Steunebrink, and
  Schmidhuber}]{greff2015lstm}
Klaus Greff, Rupesh~Kumar Srivastava, Jan Koutn{\'\i}k, Bas~R Steunebrink, and
  J{\"u}rgen Schmidhuber. 2015.
\newblock Lstm: A search space odyssey.
\newblock \emph{arXiv preprint arXiv:1503.04069}.

\bibitem[{Han et~al.(2018)Han, Cao, Lv, Lin, Liu, Sun, and Li}]{han2018openke}
Xu~Han, Shulin Cao, Xin Lv, Yankai Lin, Zhiyuan Liu, Maosong Sun, and Juanzi
  Li. 2018.
\newblock Openke: An open toolkit for knowledge embedding.
\newblock In \emph{Proceedings of the 2018 Conference on Empirical Methods in
  Natural Language Processing: System Demonstrations}, pages 139--144.

\bibitem[{Ioffe and Szegedy(2015)}]{ioffe2015batch}
Sergey Ioffe and Christian Szegedy. 2015.
\newblock Batch normalization: Accelerating deep network training by reducing
  internal covariate shift.
\newblock \emph{arXiv preprint arXiv:1502.03167}.

\bibitem[{Khot et~al.(2018)Khot, Sabharwal, and Clark}]{khot2018scitail}
Tushar Khot, Ashish Sabharwal, and Peter Clark. 2018.
\newblock Scitail: A textual entailment dataset from science question
  answering.
\newblock In \emph{Proceedings of AAAI}.

\bibitem[{Kingma and Ba(2014)}]{kingma2014adam}
Diederik Kingma and Jimmy Ba. 2014.
\newblock Adam: A method for stochastic optimization.
\newblock \emph{arXiv preprint arXiv:1412.6980}.

\bibitem[{Lample et~al.(2016)Lample, Ballesteros, Subramanian, Kawakami, and
  Dyer}]{lample2016neural}
Guillaume Lample, Miguel Ballesteros, Sandeep Subramanian, Kazuya Kawakami, and
  Chris Dyer. 2016.
\newblock Neural architectures for named entity recognition.
\newblock \emph{arXiv preprint arXiv:1603.01360}.

\bibitem[{Lan and Xu(2018)}]{lan2018neural}
Wuwei Lan and Wei Xu. 2018.
\newblock Neural network models for paraphrase identification, semantic textual
  similarity, natural language inference, and question answering.
\newblock \emph{arXiv preprint arXiv:1806.04330}.

\bibitem[{Liang(2016)}]{liang2016learning}
Percy Liang. 2016.
\newblock Learning executable semantic parsers for natural language
  understanding.
\newblock \emph{arXiv preprint arXiv:1603.06677}.

\bibitem[{Luong et~al.(2015)Luong, Pham, and Manning}]{luong2015effective}
Minh-Thang Luong, Hieu Pham, and Christopher~D Manning. 2015.
\newblock Effective approaches to attention-based neural machine translation.
\newblock \emph{arXiv preprint arXiv:1508.04025}.

\bibitem[{MacCartney and Manning(2009)}]{maccartney2009extended}
Bill MacCartney and Christopher~D Manning. 2009.
\newblock An extended model of natural logic.
\newblock In \emph{Proceedings of the eighth international conference on
  computational semantics}, pages 140--156. Association for Computational
  Linguistics.

\bibitem[{McCann et~al.(2018)McCann, Keskar, Xiong, and
  Socher}]{mccann2018natural}
Bryan McCann, Nitish~Shirish Keskar, Caiming Xiong, and Richard Socher. 2018.
\newblock The natural language decathlon: Multitask learning as question
  answering.
\newblock \emph{arXiv preprint arXiv:1806.08730}.

\bibitem[{Mitchell et~al.(2015)Mitchell, Cohen, Hruschka~Jr, Talukdar,
  Betteridge, Carlson, Mishra, Gardner, Kisiel, Krishnamurthy
  et~al.}]{mitchell2015never}
Tom~M Mitchell, William~W Cohen, Estevam~R Hruschka~Jr, Partha~Pratim Talukdar,
  Justin Betteridge, Andrew Carlson, Bhavana~Dalvi Mishra, Matthew Gardner,
  Bryan Kisiel, Jayant Krishnamurthy, et~al. 2015.
\newblock Never ending learning.
\newblock In \emph{AAAI}, pages 2302--2310.

\bibitem[{Nickel et~al.(2016)Nickel, Murphy, Tresp, and
  Gabrilovich}]{nickel2016review}
Maximilian Nickel, Kevin Murphy, Volker Tresp, and Evgeniy Gabrilovich. 2016.
\newblock A review of relational machine learning for knowledge graphs.
\newblock \emph{Proceedings of the IEEE}, 104(1):11--33.

\bibitem[{Niu et~al.(2012)Niu, Zhang, R{\'e}, and Shavlik}]{niu2012deepdive}
Feng Niu, Ce~Zhang, Christopher R{\'e}, and Jude~W Shavlik. 2012.
\newblock Deepdive: Web-scale knowledge-base construction using statistical
  learning and inference.
\newblock \emph{VLDS}, 12:25--28.

\bibitem[{Pennington et~al.(2014)Pennington, Socher, and
  Manning}]{pennington2014glove}
Jeffrey Pennington, Richard Socher, and Christopher~D Manning. 2014.
\newblock Glove: Global vectors for word representation.
\newblock In \emph{EMNLP}, volume~14, pages 1532--1543.

\bibitem[{Radford et~al.(2018)Radford, Narasimhan, Salimans, and
  Sutskever}]{radford2018improving}
Alec Radford, Karthik Narasimhan, Tim Salimans, and Ilya Sutskever. 2018.
\newblock Improving language understanding by generative pre-training.
\newblock \emph{URL https://s3-us-west-2. amazonaws.
  com/openai-assets/research-covers/languageunsupervised/language understanding
  paper. pdf}.

\bibitem[{Tai et~al.(2015)Tai, Socher, and Manning}]{tai2015improved}
Kai~Sheng Tai, Richard Socher, and Christopher~D Manning. 2015.
\newblock Improved semantic representations from tree-structured long
  short-term memory networks.
\newblock \emph{arXiv preprint arXiv:1503.00075}.

\bibitem[{Talmor and Berant(2018)}]{talmor2018web}
Alon Talmor and Jonathan Berant. 2018.
\newblock The web as a knowledge-base for answering complex questions.
\newblock \emph{arXiv preprint arXiv:1803.06643}.

\bibitem[{Wang et~al.(2015)Wang, Berant, and Liang}]{wang2015building}
Yushi Wang, Jonathan Berant, and Percy Liang. 2015.
\newblock Building a semantic parser overnight.
\newblock In \emph{Proceedings of the 53rd Annual Meeting of the Association
  for Computational Linguistics and the 7th International Joint Conference on
  Natural Language Processing (Volume 1: Long Papers)}, volume~1, pages
  1332--1342.

\bibitem[{Welbl et~al.(2018)Welbl, Stenetorp, and
  Riedel}]{welbl2018constructing}
Johannes Welbl, Pontus Stenetorp, and Sebastian Riedel. 2018.
\newblock Constructing datasets for multi-hop reading comprehension across
  documents.
\newblock \emph{Transactions of the Association of Computational Linguistics},
  6:287--302.

\bibitem[{Xiong et~al.(2017)Xiong, Hoang, and Wang}]{xiong2017deeppath}
Wenhan Xiong, Thien Hoang, and William~Yang Wang. 2017.
\newblock Deeppath: A reinforcement learning method for knowledge graph
  reasoning.
\newblock \emph{arXiv preprint arXiv:1707.06690}.

\bibitem[{Xu et~al.(2017)Xu, Liu, and Song}]{xu2017sqlnet}
Xiaojun Xu, Chang Liu, and Dawn Song. 2017.
\newblock Sqlnet: Generating structured queries from natural language without
  reinforcement learning.
\newblock \emph{arXiv preprint arXiv:1711.04436}.

\bibitem[{Yang et~al.(2018)Yang, Qi, Zhang, Bengio, Cohen, Salakhutdinov, and
  Manning}]{yang2018hotpotqa}
Zhilin Yang, Peng Qi, Saizheng Zhang, Yoshua Bengio, William~W Cohen, Ruslan
  Salakhutdinov, and Christopher~D Manning. 2018.
\newblock Hotpotqa: A dataset for diverse, explainable multi-hop question
  answering.
\newblock \emph{arXiv preprint arXiv:1809.09600}.

\bibitem[{Zhang et~al.(2018{\natexlab{a}})Zhang, Dai, Kozareva, Smola, and
  Song}]{zhang2018variational}
Yuyu Zhang, Hanjun Dai, Zornitsa Kozareva, Alexander~J Smola, and Le~Song.
  2018{\natexlab{a}}.
\newblock Variational reasoning for question answering with knowledge graph.
\newblock In \emph{Thirty-Second AAAI Conference on Artificial Intelligence}.

\bibitem[{Zhang et~al.(2018{\natexlab{b}})Zhang, Dai, Toraman, and
  Song}]{zhang2018kg}
Yuyu Zhang, Hanjun Dai, Kamil Toraman, and Le~Song. 2018{\natexlab{b}}.
\newblock Kg\^{} 2: Learning to reason science exam questions with contextual
  knowledge graph embeddings.
\newblock \emph{arXiv preprint arXiv:1805.12393}.

\bibitem[{Zhong et~al.(2017)Zhong, Xiong, and Socher}]{zhong2017seq2sql}
Victor Zhong, Caiming Xiong, and Richard Socher. 2017.
\newblock Seq2sql: Generating structured queries from natural language using
  reinforcement learning.
\newblock \emph{arXiv preprint arXiv:1709.00103}.

\bibitem[{Zhou et~al.(2018)Zhou, Huang, and Zhu}]{zhou2018interpretable}
Mantong Zhou, Minlie Huang, and Xiaoyan Zhu. 2018.
\newblock An interpretable reasoning network for multi-relation question
  answering.
\newblock \emph{arXiv preprint arXiv:1801.04726}.

\end{thebibliography}

\end{document}